\documentclass[11pt,twoside,a4paper]{article}

\usepackage{booktabs} % For formal tables
\usepackage{graphicx}
\usepackage{caption}
\usepackage{subcaption}
\usepackage[ ]{algorithm2e}

\begin{document}
\title{A Comparison of LSTM and BERT \\for Small Corpus}

\author{Aysu Ezen-Can \\ SAS Inst.}

\maketitle

\newcommand{\keywords}[1]{\textbf{\textit{Keywords:}} #1}

\begin{abstract}
Recent advancements in the NLP field showed that transfer learning helps with achieving state-of-the-art results for new tasks by tuning pre-trained models instead of starting from scratch. Transformers have made a significant improvement in creating new state-of-the-art results for many NLP tasks including but not limited to text classification, text generation, and sequence labeling. Most of these success stories were based on large datasets. In this paper we focus on a real-life scenario that scientists in academia and industry face frequently: \textit{given a small dataset, can we use a large pre-trained model like BERT and get better results than simple models?} To answer this question, we use a small dataset for intent classification collected for building chatbots and compare the performance of a simple bidirectional LSTM model with a pre-trained BERT model. Our experimental results show that bidirectional LSTM models can achieve significantly higher results than a BERT model for a small dataset and these simple models get trained in much less time than tuning the pre-trained counterparts. We conclude that the performance of a model is dependent on the task and the data, and therefore before making a model choice, these factors should be taken into consideration instead of directly choosing the most popular model. 

\end{abstract}

\section{Introduction}

Up until a couple of years ago, the natural language processing (NLP) community had been mostly training models from scratch for many different NLP tasks. Training from scratch for each task can be costly as it requires collecting a large dataset, labeling it, finding the optimal architecture, tuning parameters, and evaluating the results for the given task. Therefore, it is much more desirable to use previous knowledge gained from other tasks. The turning point for NLP came when transfer learning became possible by training a language model and using the information learned from the language model in many other NLP tasks. This is also referred to as the NLP's ImageNet moment \cite{ruder2018nlp} as ImageNet opened the doors for transfer learning in the computer vision domain and achieved new state-of-the-art results using deep learning.

In addition to transfer learning, Transformers started a new era in the NLP field. Transformers are deep learning models that can handle sequential data but they don't require sequential data to be processed in order, unlike recurrent neural networks (RNNs) \cite{transformer}. Therefore they are parallelizable, reducing the time it takes to train and enabling scientists to train on much larger datasets.

Many studies have showed that Transformers are highly successful in many NLP tasks, including summarization, translation, and classification \cite{glue}.  BERT is one of the architectures that utilizes Transformers and the model, trained in an unsupervised manner on large datasets, can be utilized in many other NLP tasks \cite{bert}. Other studies built upon BERT architecture have shown record breaking results, as shown in the GLUE Benchmark \cite{glue}. One common feature in these studies is that, even for tuning the pre-trained models, very large datasets are used. In this study, we wanted to approach this problem from a different angle: `what if we have a small dataset?' We formulated our research question around the very common real-life use case of having a small, task-specific dataset. If collecting and labeling more data is costly, and obtaining more hardware is not possible, can we still use BERT for our task? Should we forget about everything we knew about RNNs/LSTMs and completely switch to Transformers?

To the best of our knowledge, Transformers have not been compared with traditional LSTM models for task-specific small datasets. Our goal in this paper is to evaluate the use of the BERT model in a dialogue domain, where the interest for building chatbots is increasing daily. We conducted experiments for comparing BERT and LSTM in the dialogue systems domain because the need for good chatbots, expert systems and dialogue systems is high. 

\section{Related Work}

Transfer learning is the task of transferring knowledge from one task to another to reduce the effort of collecting training data and rebuilding models \cite{transferLearning}. This task is being largely adopted by the artificial intelligence community as it reuses knowledge and experience gained from one task on a completely new task, which decreases the overall time required to obtain results with good accuracy. 

For computer vision, the long-awaited transfer learning moment came with ImageNet \cite{imagenet}. In 2012, the deep neural network submitted by Krizhevsky et al.  performed 41\% better than the next best competitor \cite{imagenet}. This achievement showed the importance of both deep learning in the machine learning tasks and the importance of transfer learning. 

For NLP, the transfer learning moment did not arrive until a couple of years ago. Pennington et al. \cite{pennington2014glove} proposed a widely used vector representation for words called GloVe embeddings. However, GloVe embeddings do not utilize context while creating the word embeddings. In other words the embedding for `experiment' would be the same no matter which sentence it is used in. To address this limitation, ELMo came up with the idea of \textbf{contextualized word-embeddings} \cite{elmo} which created word embeddings using bidirectional LSTM trained with a language modeling objective. ULMFiT was also a successful model for training a neural network with language modeling objective as a precursor to fine-tuning for a specific task  \cite{ulmfit}.
 
These models were all trained on a language modeling task which enabled the use of unlabeled data at the pre-training stage. The goal in language modeling is to predict the next word based on the previous words. 
BERT is different from ELMo and ULMFiT because it uses a masked language modeling approach. BERT addresses the limitations in prior work by taking the contexts of both the previous and next words into account instead of just looking to the next word for context. In the masked language modeling approach, words in a sentence are randomly erased and replaced with a special token, and a Transformer is used to generate a prediction for the masked word based on the unmasked words surrounding it.

With the masked language modeling objective, BERT achieved record breaking results in many NLP tasks as shown in GLUE benchmark \cite{glue}. Many other Transformer architectures followed BERT, such as RoBERTa \cite{roberta}, DistillBERT \cite{distilbert}, OpenAI Transformer \cite{openai},and XLNet \cite{xlnet}, achieving incremental results.

Having recognized the record breaking success of Transformers, our goal in this paper is to compare the performance of Transformers with traditional bidirectional LSTM models for a small dataset. GLUE benchmark showed that utilizing BERT-like models for large datasets is successful already. This paper will show a comparison for a task-specific small dataset.

\section{Methodology}

\begin{figure}
\centering
\includegraphics[width=0.8\textwidth]{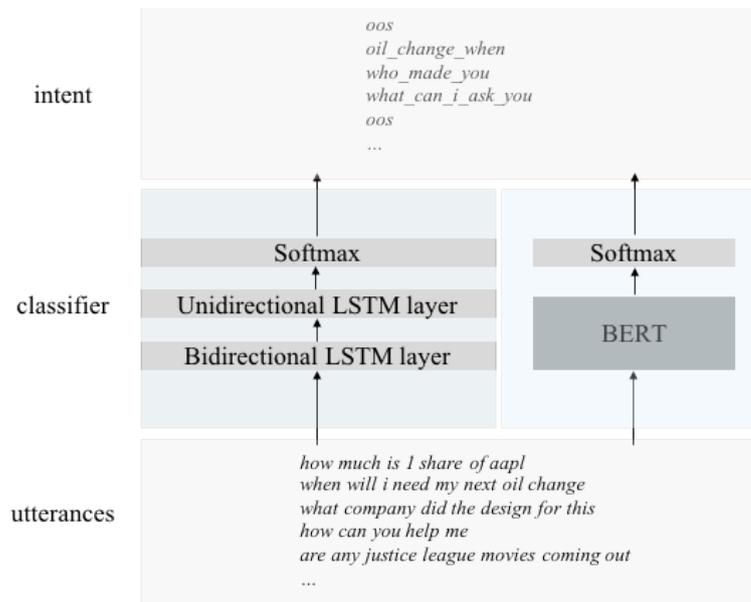}
\caption{Pipeline showing the inputs and outputs of classifiers (LSTM and BERT).
  }
 \label{fig:pipeline}
\end{figure}

To compare BERT and LSTM, we chose a text classification task and trained both BERT and LSTM on the same training sets, evaluated with the same validation and test sets. Because our goal is to evaluate these models on small datasets, we randomly split the datasets into smaller versions by taking $X$ percent of the data where $X \in \{25,40,50,60,70,80,90\}$. Figure \ref{fig:pipeline} depicts the pipeline comparing LSTM and BERT for the intent classification task. As can be seen in the figure, the classifiers take utterances (chatbot interactions) in text format and predict the nominal outcome (intent). Each utterance has an intent label in the dataset and the goal of the classifiers is to predict the intent of an utterance as correctly as possible.

In order to compare different LSTM models, we experimented with six different architectures by varying the number of neurons in the LSTM layers and the number of bidirectional layers. We experimented with three LSTM models with 50 neurons in each of the LSTM layers, as well as three LSTM models with 100 neurons in each LSTM layer. 

\section{Experiments}
In this section, we first discuss the corpora used in this study and then provide experimental results. 

\subsection{Corpora}

We chose a small dataset for our comparisons \cite{larson2019evaluation}. Table \ref{data} shows the number of utterances in training, validation and test sets. This dataset contains utterances collected for building a chatbot. It has 150 intent classes with 100 training observations from each class. For each intent, 20 validation and 30 test queries are provided. There are also out-of-scope queries that do not fall under any of the 150 intent classes. The goal of this dataset is to challenge the dialogue modeling field to also focus on out-of-scope utterances. The goal of the model is to label them out-of-scope in order for the chatbot not to proceed with unintended system utterances. However, the challenge comes from the fact that the dataset is very skewed; the number of out-of-scope utterances is very low compared to in-scope utterances and the number of classes are high compared to the number of observations available for each intent. Only a total of 1,200 out-of-scope utterances exist in the dataset, with 100 of them in the training set, 100 of them in the validation set and the remaining 1,000 out-of-scope utterances in the test set. It is important to note that we did not want to change the ordering or distribution of the data to be able to make a fair comparison with the paper that introduced this dataset \cite{larson2019evaluation}.

\begin{center}
\begin{table}[t]
\centering
\begin{tabular}{ c c } 
\hline
Dataset &  Number of utterances  \\
 \hline
Training  & 15,101  \\ 
Validation  & 3,101 \\ 
Test & 5,501\\ 
\hline
\end{tabular}
\caption{\label{data} Number of utterances in each set.}
\end{table}
\end{center}

An excerpt from the corpus can be found in Table \ref{data}. In the original paper, Larson et al. compare several different models including support vector machines, convolutional neural networks and BERT. However, they do not compare LSTM and don't report overall accuracy \cite{larson2019evaluation}. Larson et al. conclude that BERT performs best when accuracy is calculated for the in-scope utterances only and all models suffer when there are out-of-scope utterances. In this paper, we compare LSTM and BERT and provide overall accuracy metric which includes both the 150 in-scope utterances and the out-of-scope utterances. 

\begin{center}
\begin{table}[t]
\centering
\begin{tabular}{ l r } 
\hline
\textbf{Utterance} &  \textbf{Intent}  \\
 \hline
how do you say hi in french  & translate  \\ 
in england how do they say subway  & translate  \\ 
in england how do they say subway  & translate  \\ 
i was at whole foods try…nd my card got declined  & card$\_$declined  \\ 
when's the next time i have to pay the insurance & bill$\_$due \\
what about the calories for this chicken salad & calories \\ 
my card got snapped in half & damaged$\_$card \\
how much is an overdraft fee for bank & \textbf{oos} \\
how's the lo mein rated at hun lee's & restaurant$\_$reviews \\
can you tell me if eating at outback's any good & restaurant$\_$reviews \\
what peruvian dish should i make & meal$\_$suggestion \\
how much did i earn in income only last year & income \\
how much money does radiohead earn a year & \textbf{oos} \\
\hline
\end{tabular}
\caption{\label{data} Excerpt from the corpus}
\end{table}
\end{center}

\begin{figure}
\centering
\includegraphics[width=0.6\textwidth]{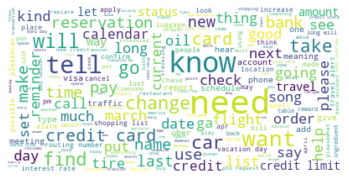}
\caption{Word cloud for the training set.
  }
 \label{fig:training_word_cloud}
\end{figure}

As can be seen in the word cloud (Figure \ref{fig:training_word_cloud}), there are words from many different topics in the training set and none of the words' frequencies dominate others. This shows how challenging this dataset is, as it is not enough for a model to learn a specific set of words that go together nor possible to memorize any patterns. 

\subsection{Experimental Results}

\begin{figure}
\centering
\includegraphics[width=1\textwidth]{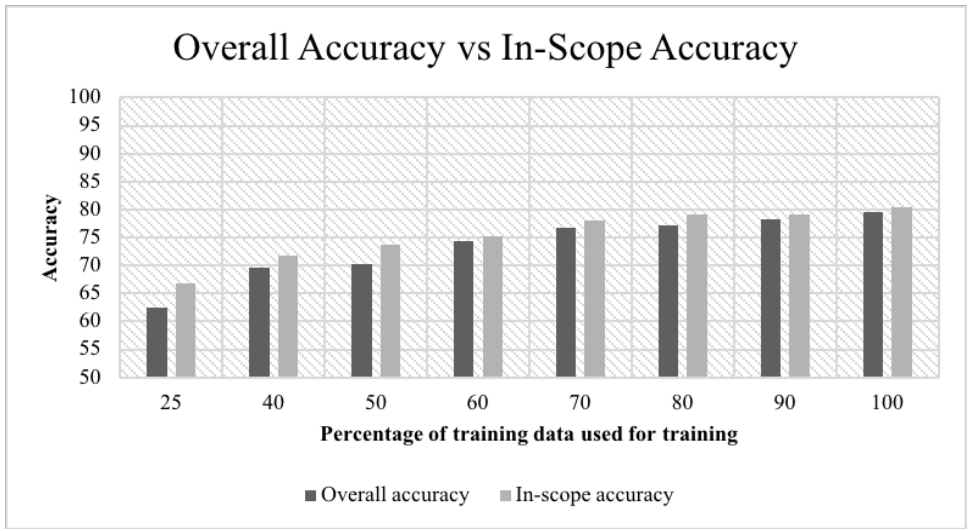}
\caption{Overall accuracy vs. in-scope accuracy
  }
 \label{fig:overall_vs_inscopeAccuracy}
\end{figure}

\begin{figure}
\centering
\includegraphics[width=0.7\textwidth]{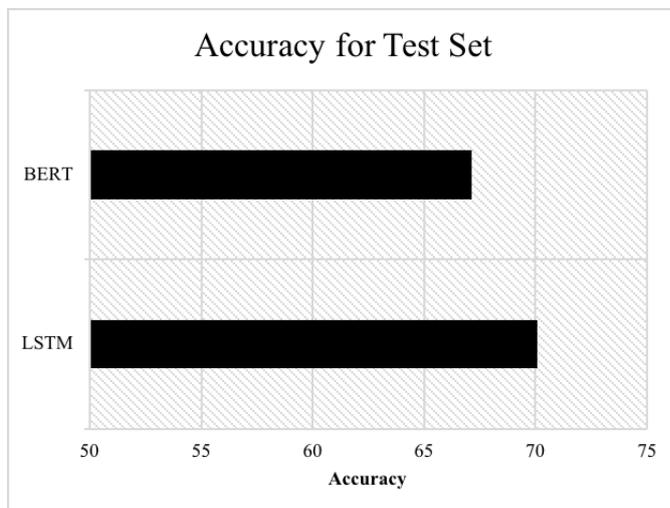}
\caption{Accuracy comparison for BERT and LSTM with the test set.
  }
 \label{fig:test_accuracy}
\end{figure}

In this section, we present experimental results comparing LSTM and BERT for the intent classification task. In addition to the overall accuracy, we also report in-scope accuracy as done by \cite{larson2019evaluation}. In-scope accuracy is calculated using only in-scope utterances before calculating the accuracy metric. However, we believe that overall accuracy is a better metric for measuring the overall performance of the model as it does not remove the challenging utterances from the result and therefore is more realistic. For comparing overall accuracy and in-scope accuracy, Figure \ref{fig:overall_vs_inscopeAccuracy} shows them side by side for different versions of the dataset. As shown in the figure, in all 8 partitions of the data, in-scope accuracy results are higher than overall accuracy results. We therefore report only overall accuracy for comparing BERT and LSTM (see Figure \ref{fig:test_accuracy}).

\textbf{\textit{Model parameters.}} For tuning the BERT model, a learning rate of 2e-5 was used. While training the LSTM model, a learning rate of 0.01 was used with Adam optimizer. While BERT utilizes its own embeddings, for the LSTM model we used Glove embeddings \cite{pennington2014glove}. 

\textbf{\textit{LSTM architectures.}} We experimented with different LSTM architectures, as can be seen in Table \ref{architectures}. Interestingly, the simplest LSTM model performed the best in terms of both overall accuracy (both in-scope and out-of-scope accuracy) and in-scope accuracy (removing the out-of-scope utterances to calculate accuracy).

\textbf{\textit{Toolkit.}} Throughout our studies, we used SAS Deep Learning. We used Bert base uncased from HuggingFace repository as the pre-trained model and used SAS Deep Learning to fine-tune the model. LSTM models were built from scratch.

\begin{figure}
\centering
\includegraphics[width=1\textwidth]{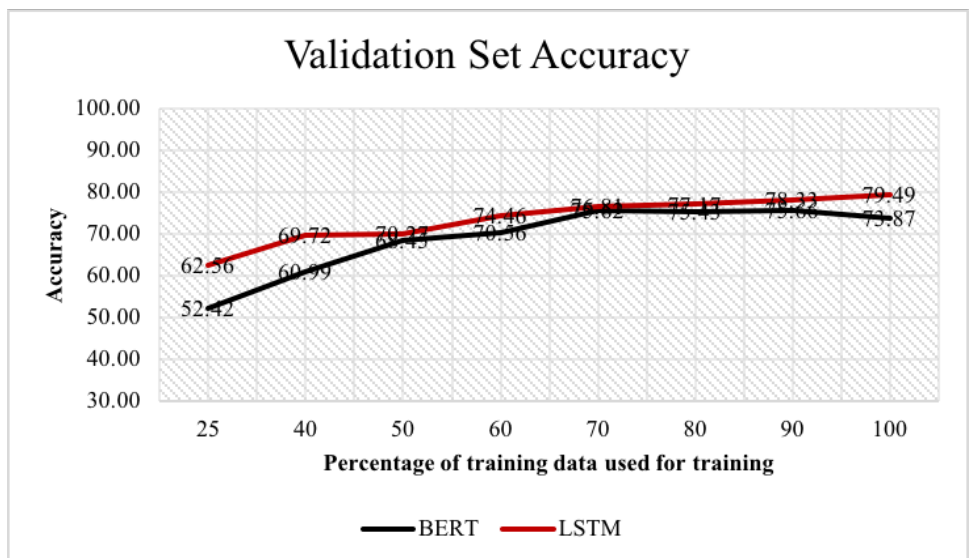}
\caption{Overall accuracy for all classes scored on validation set.
  }
 \label{fig:validation_accuracy}
\end{figure}

\begin{center}
\begin{table}[t]
\centering
\begin{tabular}{ c c c c } 
\hline
\textbf{LSTM}  & \# \textbf{of neurons} & \textbf{overall} & \textbf{in-scope}  \\
\textbf{architecture}&&\textbf{accuracy}&\textbf{accuracy} \\
 \hline
1 bidirectional + 1 unidirectional &50& \textbf{70.08} & \textbf{69.65} \\
2 bidirectional + 1 unidirectional &50& 63.22 & 63.72 \\
3 bidirectional + 1 unidirectional &50& 49.86 & 51.41 \\
1 bidirectional + 1 unidirectional &100& 66.88 & 67.12 \\
2 bidirectional + 1 unidirectional &100& 49.61 &  50.91\\
3 bidirectional + 1 unidirectional & 100&7.05 &  7.71\\
\hline
\end{tabular}
\caption{\label{architectures} Evaluation criteria with different LSTM architectures.}
\end{table}
\end{center}

\textbf{\textit{Validation set.}} In this study, the validation set is only used for determining which architecture is best, so that it can be used for final scoring with the test data. Models were not trained or tuned using the validation data. Figure \ref{fig:validation_accuracy} compares BERT and the simple LSTM architecture (1 bidirectional layer and 1 unidirectional layer with 50 neurons in each layer) for different data sizes (from 25\% of training data used to 100\% of the data used). The experimental results show that LSTM outperforms BERT in every data partition. Paired 2-tailed t-test show that the results between LSTM (1 bidirectional layer + unidirectional layer) and BERT are statistically significant ($p < 0.008$). One interesting finding is that, the difference in terms of accuracy between LSTM and BERT is much more when the dataset is small than when the dataset is larger (16.21$\%$ relative difference with 25$\%$ of the dataset versus 2.25$\%$ relative difference with 80$\%$ of the dataset). This finding shows that with small datasets, simple models such as LSTM can perform better where complex models such as BERT may overfit.

\textbf{\textit{Test set.}} Through our analyses, we found that the simplest LSTM model performed best with the validation dataset, therefore we chose the simplest LSTM model to compare with BERT in the test set. Note that the test set has a much larger set of out-of-scope utterances and therefore it is expected for accuracies to be lower than validation set. In real-time systems, the chatbot is challenged by unseen out-of-scope utterances when users are interacting with the system therefore we expect dialogue understanding models to be robust to these types of utterances. This is the reason why the test set has more out-of-scope utterances -- to make sure the model have not had much experience with out-of-scope utterances but still can work with them in real time. With the test set, in-score accuracy for the LSTM model was 69.65$\%$ and overall accuracy was 70.08$\%$ whereas BERT model achieved 67.15$\%$ accuracy. Figure \ref{fig:test_accuracy} shows the comparison between BERT and LSTM. As can be seen in the test set comparisons as well as validation set comparisons, LSTM performs with higher accuracy than BERT for this small dataset.

\section{Conclusion}

BERT architecture achieved a breakthrough in the NLP field in many tasks and made it possible to utilize transfer learning. This approach opened the doors for training with large unlabeled datasets in an unsupervised manner and modifying the last layers of the model to adapt it to particular tasks. Then with tuning the parameters, many scientists achieved new state-of-the-art results for many different datasets \cite{glue}. Many of these studies utilized large datasets for tuning the models and scoring. In this study, we approached these models from a different angle by focusing on a small dataset. 

We formulated our research question to be about comparing LSTM models with BERT for task-specific small datasets. To that end, we chose an intent classification task with a small dataset collected for building chatbots. Collecting data with human studies is costly and time consuming, as is manual labeling. Therefore, chatbots is a suitable domain to choose for a small dataset, as the chatbot interactions for a dataset need to be collected via human studies rather than being able to use synthetic data. 

We experimented with different LSTM architectures and found that the simplest LSTM architecture we tried worked best for this dataset. When compared to BERT, LSTM statistically significantly performed with higher accuracy in both validation data and test data. In addition, the experimental results showed that for smaller datasets, BERT overfits more than simple LSTM architecture.

To sum up, this study is by no means to undermine the success of BERT. However, we compared LSTM and BERT from a small dataset perspective and experimental results showed that LSTM could have higher accuracy with less time to build and tune models for certain datasets, such as the intent classification data that we focused on. Therefore, it is important to analyze the dataset and research/business needs first before making a decision on which models to use. 

\newpage
\bibliographystyle{plain}
\bibliography{sample-bibliography}

\end{document}